\documentclass[runningheads]{llncs}
\usepackage{graphicx}
\usepackage{adjustbox}
\usepackage{multicol}
\usepackage{multirow}
\usepackage{amsmath}
\usepackage{hyperref}

\usepackage{xcolor}

\begin{document}
\title{Deepfake Style Transfer Mixture: a First Forensic Ballistics Study on Synthetic Images}
\titlerunning{A First Forensic Ballistics Study on Synthetic Images}
%
\author{Luca Guarnera\inst{1,2}\orcidID{0000-0001-8315-351X} \and
Oliver Giudice\inst{1,3}\orcidID{0000-0002-8343-2049} \and
Sebastiano Battiato\inst{1,2}\orcidID{0000-0001-6127-2470}}

\authorrunning{L. Guarnera et al.}

\institute{Department of Mathematics and Computer Science, University of Catania, Italy \and
iCTLab Spinoff of University of Catania, Italy\and
Applied Research Team, IT dept., Banca d'Italia, Italy \\ 
\email{luca.guarnera@unict.it}, \email{\{giudice,battiato\}@dmi.unict.it}}
\maketitle              
\begin{abstract}
Most recent style-transfer techniques based on generative architectures are able to obtain synthetic multimedia contents, or commonly called deepfakes, with almost no artifacts. Researchers already demonstrated that synthetic images contain patterns that can determine not only if it is a deepfake but also the generative architecture employed to create the image data itself. These traces can be exploited to study problems that have never been addressed in the context of deepfakes. To this aim, in this paper a first approach to investigate the image ballistics on deepfake images subject to style-transfer manipulations is proposed. Specifically, this paper describes a study on detecting how many times a digital image has been processed by a generative architecture for style transfer. Moreover, in order to address and study accurately forensic ballistics on deepfake images, some mathematical properties of style-transfer operations were investigated.  

\keywords{Image Ballistics  \and Deepfake \and Multimedia Forensics.}
\end{abstract}
\section{Introduction}

Advances in Deep learning algorithms and specifically the introduction of Generative Adversarial Networks (GAN)~\cite{goodfellow2014generative} architectures, enabled the creation and widespread of extremely refined techniques able to \emph{manipulate} digital data, alter it or create contents from scratch. These algorithms achieved surprisingly realistic results leading to the birth of the Deepfake phenomenon: multimedia contents synthetically modified or created through machine learning techniques. Deepfakes, or also called synthetic multimedia content, could be employed to manipulate or even generate realistic places, animals, objects and human beings. In this paper only the most dangerous deepfake images of people's faces were taken into account.
Specifically, manipulations on people's faces could be categorized into four main groups:
\begin{itemize}
\item \textbf{Entire face synthesis}: creates entire non-existent face images (\cite{karras2019style,karras2020analyzing,karras2020training}). 
\item \textbf{Attribute manipulation}: also known as \textit{face editing} or \textit{facial retouching}, modifies facial attributes~\cite{choi2018stargan,choi2020stargan} such as hair color, gender, age. 
\item \textbf{Identity swap}: replaces a person's face in a video with another person's face~\footnote{\url{https://github.com/deepfakes/faceswap}}). 
\item \textbf{Expression swap}: also known as \textit{face reenactment}~\cite{thies2016face2face,thies2019deferred}, modifies the person's facial expression.
\end{itemize}

\begin{figure}[t!]
    \includegraphics[width=\linewidth]{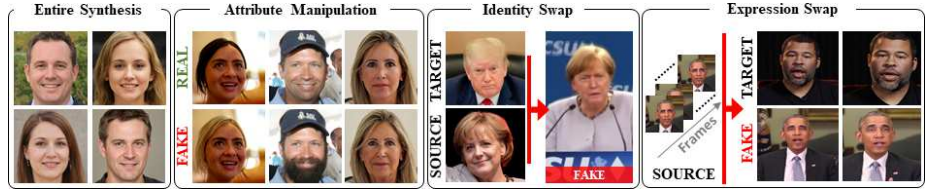}
    \caption{Deepfake manipulation categories.
 }
    \label{fig:manipulation}
\end{figure}

Figure~\ref{fig:manipulation} shows an example of each kind of manipulations.

Unfortunately, these types of manipulations are often used for malicious purposes such as industrial espionage attacks and threats to individuals having their face placed in a porn video. To counteract the illicit use of this powerful technology, forensic researchers have created in the last years several deepfake detection algorithms able to solve the \textit{Real} Vs \textit{Deepfake} classification task. 
State-of-the-art methods have demonstrated that Convolutional Neural Networks (CNNs)~\cite{xuan2019generalization,hsu2020deep,gandhi2020adversarial,li2020face,wang2020cnn} and analytical approaches~\cite{guarnera2020deepfake,guarnera2020fighting,mccloskey2018detecting,giudice2021fighting} could be employed to extract a unique fingerprint on synthetic content to define not only whether the digital data are deepfakes but also to determine the GAN architecture that was used in the creation phase.
These unique patterns can be extracted from deepfake images to analyze and study them from the perspective of the science of Image Ballistics~\cite{farid2008digital,giudice2017classification,piva2013overview,battiato2016multimedia}, in order to reconstruct the history of the data itself (a study that has never been addressed in the world of deepfakes). A first basic approach is proposed in this paper by defining whether a deepfake images is subject to single or double deepfake manipulation operations. In our experiments, only the Style-Transfer category as deepfake manipulation type was taken into account. Several mathematical properties of style-transfer operations were also analysed and the obtained results could give extremely interesting hints to further study the Forensic Ballistics task on Deepfake images with techniques similar to double quantization detection~\cite{giudice20191}.

The remainder of this paper is organized as follows: Section~\ref{sec:related} presents some state-of-the-art methods of Deepfakes creation in order to understand the process of creating synthetic digital content. The proposed approach and experimental results 
are described in Section~\ref{sec:fb}. In Section~\ref{sec:mat} some mathematical properties of the Style-Transfer operation are demonstrated. Finally, Section~\ref{sec:conclusion} concludes the paper.

\section{State of the art}
\label{sec:related}

An overview on Media forensics with particular focus on Deepfakes has been proposed in~\cite{verdoliva2020media,guarnera2020preliminary,zhang2022deepfake}.
Deepfakes are generally created by Generative Adversarial Networks (GANs) firstly introduced by Goodfellow et al.~\cite{goodfellow2014generative}. Authors proposed a new framework for estimating generative models via an adversarial mode in which two models train simultaneously: a generative model $G$, that captures the data distribution, and a discriminative model $D$, able to estimate the probability that a sample comes from the training data rather than from $G$. The training procedure for $G$ is to maximize the probability of $D$ making an error, resulting in a min-max two-player game.

The best results in the deepfake creation process were achieved by StyleGAN~\cite {karras2019style} and StyleGAN2~\cite{karras2020analyzing}, two \textit{entire face synthesis} algorithms. StyleGAN~\cite{karras2019style}, while being able to create realistic pseudo-portraits of high quality, 
small artifacts could reveal the fakeness of the generated images.  To correct these imperfections, Karras et al. made some improvements to the StyleGAN generator (including re-designed normalization, multi-resolution, and regularization methods) by proposing StyleGAN2~\cite{karras2020analyzing}.
Face attribute manipulation use image-to-image translation techniques in order to address the style diversity~\cite{almahairi2018augmented,huang2018multimodal,mao2019mode,lee2018diverse,na2019miso}. The main limitation of these methods is that they are not scalable to the increasing number of domains (e.g., it is possible to train a model to change only the air color, only the facial expression, etc.).  
To solve this problem, Choi et al. proposed StarGAN~\cite{choi2018stargan}, a framework that can perform image-to-image translations on multiple domains using a single model. Given a random label as input, such as hair color, facial expression, etc., StarGAN is able to perform an image-to-image translation operation. 
This framework has a limitation: it does not capture the multi-modal nature of data distribution. In other words, this architecture learns a deterministic mapping of each domain, i.e. given an image and a fixed label as input, the generator will produce the same output for each domain. This mainly depends on the fact that each domain is defined by a predetermined label. To solve this limitation, Choi et al. recently proposed StarGAN v2~\cite{choi2020stargan}. The authors changed the domain label to domain-specific style code by introducing two modules: (a) a mapping network to learn how to transform random Gaussian noise into a style code, and (b) a style encoder to learn how to extract the style code from a given reference image. 

He et al.~\cite{he2019attgan} proposed a new technique called AttGAN in which an attribute classification constraint is applied to the generated image, in order to guarantee only the correct modifications of the desired attributes. 
Achieved results showed that AttGAN exceeds the state of the art on the realistic modification of facial attributes. 

An interesting style transfer approach was proposed by Cho et al.~\cite{cho2019image}, with a framework called ``group-wise deep whitening-and coloring method" (GDWCT) for a better styling capacity. 
GDWCT has been compared with various cutting-edge methods in image translation and style transfer improving not only computational efficiency but also quality of generated images.



\section{Forensic Ballistics on Deepfake Images}
\label{sec:fb}

\begin{figure}[t!]
    \includegraphics[width=\linewidth]{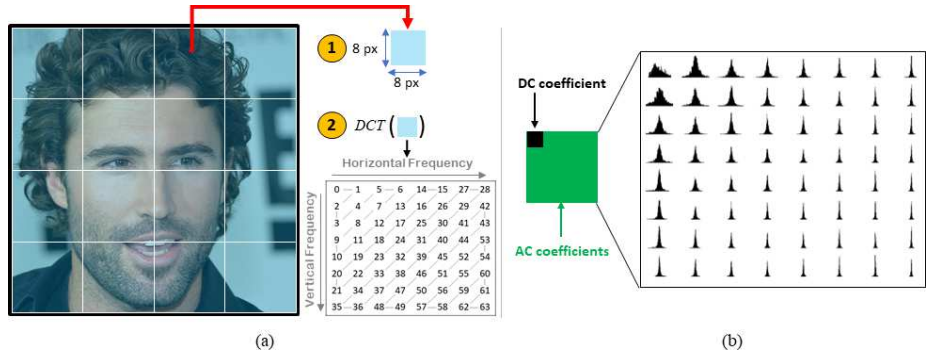}
    \caption{(a-1) The input image $I$ is divided into non-overlapping blocks of size $8 \times 8$. The DCT (a-2) is applied to each block and the DC/AC coefficients are obtained. (b) The DC histogram and the AC histograms are modeled, as demonstrated in~\cite{lam2000mathematical}, by considering all the coefficients of each block: e.g., the histogram of the AC coefficient of position 1 is obtained by modeling (following the Laplacian distribution) all the coefficients of position 1 of each block. 
 }
    \label{fig:DCT}
\end{figure}

In order to define whether a multimedia content has undergone one or two deepfake manipulation operations belonging to the style-transfer category, we considered two approaches: 
\begin{itemize}
    \item A first analytical approach (Method 1) in which the DCT coefficients were analyzed to study whether there are discriminative frequency statistics.
    \item Given the potential of deep neural networks to solve different deepafake classification tasks, our second method (Method 2) relies on the use of a basic encoder (RESNET-18).

\end{itemize}
\paragraph{\textbf{Method 1:}} \label{P1} Let $I$ be a digital image. Following the JPEG pipeline, $I$ is divided into non-overlapping blocks of size $8 \times 8$ (Figure~\ref{fig:DCT}(a-1)). The Discrete Cosine Transform (DCT) is then applied to each block obtaining 
the 64 DCT coefficient.
They are sorted into a zig-zag order starting from the top-left element to the bottom right (Figure~\ref{fig:DCT}(a-2)). The DCT coefficient at position 0 is called DC and represents the average value of pixels in the block. All others coefficients namely AC, corresponds to specific bands of frequencies. By applying evidence reported in~\cite{lam2000mathematical}, the DC coefficient can be modelled with a Gaussian distribution while the AC coefficients were demonstrated to follow a zero-centred Laplacian distribution (Figure~\ref{fig:DCT}(b)) described by:
\begin{equation}
	\label{eq:laplacian}
	P(x) =  {1\over2\beta}exp\Biggl({-|x-\mu| \over \beta} \Biggl)
\end{equation}
with $\mu=0$ and $\beta = \sigma/\sqrt{2}$ is the scale parameter where $\sigma$ corresponds to the standard deviation of the AC coefficient distributions. 
The $\beta$ values define the amplitude of the Laplacian graph  that characterizes the AC parameters. 
It is possible to analyze all $\beta$ values to determine if they can be discriminative to solve the proposed task.
For this reason, the following feature vector was obtained for each involved image:
\begin{equation}
    \vec \beta = \{\beta_{1}, \beta_{2}, \dots, \beta_{N}\} 
\end{equation}

with $N = 63$. Note that $\beta_{1}$ through $\beta_{N}$ represent only the $\beta$ values extracted from the 63 AC coefficients: the DC coefficient is excluded because it represent, as described above, only the average value of the pixels in the block. 
The $\vec\beta$ feature vectors, extracted for each involved images, were used to train the following standard classifiers: k-NN (with $k=\{1, 3, 5, 7, 11, 13, 15\}$), SVM (with linear, poly, rbf and sigmoid kernels), Linear Discriminant Analysis (LDA), Decision-Tree, Random Forest, GBoost.

\paragraph{\textbf{Method 2:}} \label{P2} 
The Pytorch implementation of RESNET-18 was used starting from the pre-trained model trained on ImageNet~\footnote{\url{https://pytorch.org/vision/stable/models.html}}. A fully-connected layer with an output size of $2$, followed by a SoftMax, were added to the last layer of the Resnet-18 in order to be trained on the specific task. The network was set considering the following parameters: $batch-size = 30$; $learning-rate = 0.0001$; $number-of-classes = 2$; $epochs = 100$; $criterion = CrossEntropyLoss$; $optimizer = SGD$ with $momentum = 0.9$.

\paragraph{\textbf{Creating datasets for both methods:}} 
Let $S(s,t) = s \bigoplus t$ be a generic style transfer architecture. The function $S$ takes as input a source $s$ and a target $t$. The style transfer operation (which we defined with the symbol $\bigoplus$) is performed between $s$ and $t$: attributes (such as hair color, facial expression) are captured by $t$ to be transferred to $s$.
So, for our task, we define two classes:

\begin{itemize}
    \item \textit{Deepfake-2} representing all images where a style transfer operation is applied once (the 2 in the Deepfake-2 class label refers to the input number: the source and the target). Thus, we collected a set of images $D_1 = S(s_1,t_1)$, where $s_1$ and $t_1$ represent two different dataset; 
    \item \textit{Deepfake-3} representing all images where a style transfer operation is applied twice. 
    Thus, we have collected a set of images $D_2 = S(D_1,t_2)$, where $D_1$ is the source representing the output of the first style transfer operation and $t_2 \neq t_1$, $t_2 \neq s_1$.
\end{itemize}

\subsection{Experiment Results}

\begin{figure}[t!]
\centering
    \includegraphics[width=0.8\linewidth]{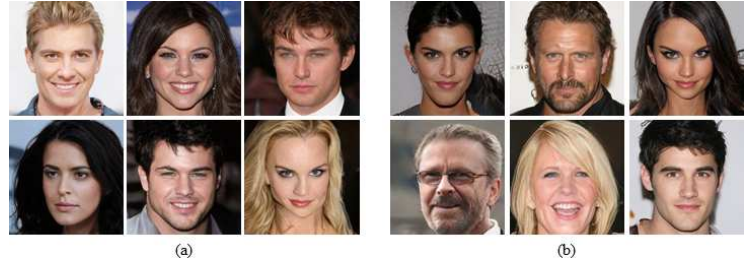}
    \caption{Examples of images generated by StarGAN-V2 engine~\cite{choi2020stargan}. (a) images in which a style transfer operation was applied once; (b) images in which a style transfer operation was applied twice.
 }
    \label{fig:starg}
\end{figure}


For our experiments, we collected images generated by StarGAN-V2~\cite{choi2020stargan}, because it is the best style transfer architecture capable of obtaining deepfake images with high quality detail and resolution.
Figure~\ref{fig:starg} shows examples of StarGAN-V2 deepfake images related to the two classes.
For the methods described above, $1200$ images were considered as the train set and $200$ images as the test set for each class. So in total, our train set consists of $2400$ images and the test set consists of $400$ images.

\begin{figure}[t!]
    \includegraphics[width=\linewidth]{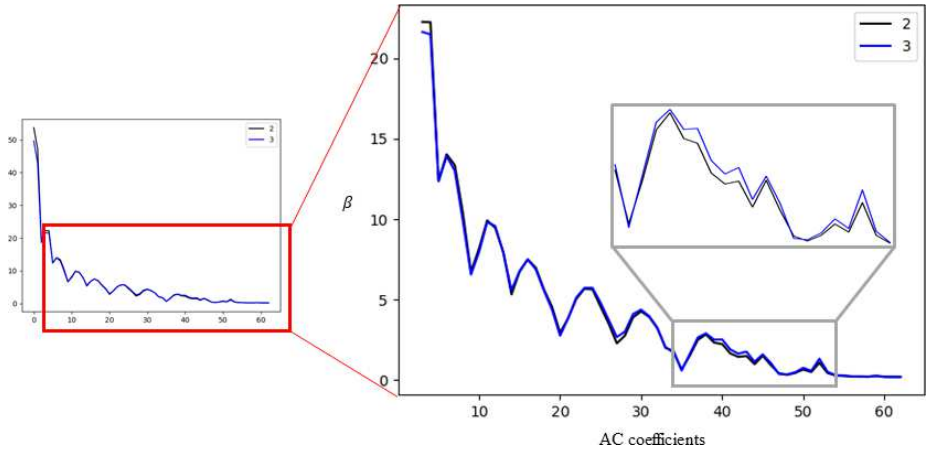}
    \caption{The average $\beta$ values for each $i$-th AC coefficient of the involved dataset are reported. Label ``2" and ``3" represents the trend of the AC coefficients related to the Deepfake-2 and Deepfake-3 class images. The abscissa axis are the 64 coefficients of the $8 \times 8$ block, while the ordinate axis are the average of the $\beta$ values.
 }
    \label{fig:Accoeff}
\end{figure}

Regarding Method 1, Figure~\ref{fig:Accoeff} shows how the feature vector $\vec\beta$ extracted from all the involved images can to be discriminative. In detail, we plot the average $\beta$ values for each $i$-th AC coefficient of the train set images. It can be seen that some $\beta$ statistics (albeit slightly) appear to be discriminative in solving the proposed task. 
Table~\ref{tab:classres} shows the classification results obtained by training the standard classifiers (listed above) considering $\vec\beta$ feature vectors as input.
As a first approach in forensic ballistics on deepfake images, we obtained at best a classification accuracy value of $81\%$ using the Random Forest classifier.


\begin{table}[t!]

\centering
\caption{Classification results of the proposed analytical method.}\label{tab:classres}
\begin{tabular}{ccc|c|c|c|c|}
\cline{4-7}
\multicolumn{2}{c}{\textbf{Classifiers}}                                                                      & \textbf{Classes} & \textbf{Precision} & \textbf{Recall} & \textbf{F1-score} & \textbf{Accuracy (\%)} \\ \hline
\multicolumn{1}{|c}{\multirow{12}{*}{\textbf{k-NN}}} & \multicolumn{1}{c|}{\multirow{2}{*}{\textbf{k = 3}}}   & Deepfake-2       & 0.78               & 0.69            & 0.73              & \multirow{2}{*}{74\%}  \\ \cline{3-6}
\multicolumn{1}{|c}{}                                & \multicolumn{1}{c|}{}                                  & Deepfake-3       & 0.72               & 0.79            & 0.75              &                        \\ \cline{2-7} 
\multicolumn{1}{|c}{}                                & \multicolumn{1}{c|}{\multirow{2}{*}{\textbf{k = 5}}}   & Deepfake-2       & 0.79               & 0.67            & 0.73              & \multirow{2}{*}{75\%}  \\ \cline{3-6}
\multicolumn{1}{|c}{}                                & \multicolumn{1}{c|}{}                                  & Deepfake-3       & 0.71               & 0.82            & 0.76              &                        \\ \cline{2-7} 
\multicolumn{1}{|c}{}                                & \multicolumn{1}{c|}{\multirow{2}{*}{\textbf{k = 7}}}   & Deepfake-2       & 0.78               & 0.66            & 0.71              & \multirow{2}{*}{73\%}  \\ \cline{3-6}
\multicolumn{1}{|c}{}                                & \multicolumn{1}{c|}{}                                  & Deepfake-3       & 0.70               & 0.81            & 0.75              &                        \\ \cline{2-7} 
\multicolumn{1}{|c}{}                                & \multicolumn{1}{c|}{\multirow{2}{*}{\textbf{k = 11}}}  & Deepfake-2       & 0.79               & 0.67            & 0.72              & \multirow{2}{*}{74\%}  \\ \cline{3-6}
\multicolumn{1}{|c}{}                                & \multicolumn{1}{c|}{}                                  & Deepfake-3       & 0.70               & 0.82            & 0.76              &                        \\ \cline{2-7} 
\multicolumn{1}{|c}{}                                & \multicolumn{1}{c|}{\multirow{2}{*}{\textbf{k = 13}}}  & Deepfake-2       & 0.79               & 0.65            & 0.71              & \multirow{2}{*}{74\%}  \\ \cline{3-6}
\multicolumn{1}{|c}{}                                & \multicolumn{1}{c|}{}                                  & Deepfake-3       & 0.70               & 0.82            & 0.75              &                        \\ \cline{2-7} 
\multicolumn{1}{|c}{}                                & \multicolumn{1}{c|}{\multirow{2}{*}{\textbf{k = 15}}}  & Deepfake-2       & 0.79               & 0.65            & 0.71              & \multirow{2}{*}{73\%}  \\ \cline{3-6}
\multicolumn{1}{|c}{}                                & \multicolumn{1}{c|}{}                                  & Deepfake-3       & 0.69               & 0.82            & 0.75              &                        \\ \hline
\multicolumn{1}{|c}{\multirow{8}{*}{\textbf{SVM}}}   & \multicolumn{1}{c|}{\multirow{2}{*}{\textbf{linear}}}  & Deepfake-2       & 0.78               & 0.65            & 0.71              & \multirow{2}{*}{73\%}  \\ \cline{3-6}
\multicolumn{1}{|c}{}                                & \multicolumn{1}{c|}{}                                  & Deepfake-3       & 0.69               & 0.81            & 0.74              &                        \\ \cline{2-7} 
\multicolumn{1}{|c}{}                                & \multicolumn{1}{c|}{\multirow{2}{*}{\textbf{Poly}}}    & Deepfake-2       & 0.71               & 0.41            & 0.52              & \multirow{2}{*}{62\%}  \\ \cline{3-6}
\multicolumn{1}{|c}{}                                & \multicolumn{1}{c|}{}                                  & Deepfake-3       & 0.58               & 0.83            & 0.68              &                        \\ \cline{2-7} 
\multicolumn{1}{|c}{}                                & \multicolumn{1}{c|}{\multirow{2}{*}{\textbf{rbf}}}     & Deepfake-2       & 0.69               & 0.41            & 0.51              & \multirow{2}{*}{61\%}  \\ \cline{3-6}
\multicolumn{1}{|c}{}                                & \multicolumn{1}{c|}{}                                  & Deepfake-3       & 0.57               & 0.81            & 0.67              &                        \\ \cline{2-7} 
\multicolumn{1}{|c}{}                                & \multicolumn{1}{c|}{\multirow{2}{*}{\textbf{Sigmoid}}} & Deepfake-2       & 0.66               & 0.51            & 0.57              & \multirow{2}{*}{62\%}  \\ \cline{3-6}
\multicolumn{1}{|c}{}                                & \multicolumn{1}{c|}{}                                  & Deepfake-3       & 0.59               & 0.73            & 0.65              &                        \\ \hline
\multicolumn{2}{|c|}{\multirow{2}{*}{\textbf{LDA}}}                                                           & Deepfake-2       & 0.77               & 0.67            & 0.72              & \multirow{2}{*}{73\%}  \\ \cline{3-6}
\multicolumn{2}{|c|}{}                                                                                        & Deepfake-3       & 0.70               & 0.79            & 0.75              &                        \\ \hline
\multicolumn{2}{|c|}{\multirow{2}{*}{\textbf{Decision-Tree}}}                                                 & Deepfake-2       & 0.76               & 0.72            & 0.74              & \multirow{2}{*}{74\%}  \\ \cline{3-6}
\multicolumn{2}{|c|}{}                                                                                        & Deepfake-3       & 0.73               & 0.76            & 0.74              &                        \\ \hline
\multicolumn{2}{|c|}{\multirow{2}{*}{\textbf{Random Forest}}}                                                 & Deepfake-2       & 0.85               & 0.76            & 0.80              & \multirow{2}{*}{81\%}  \\ \cline{3-6}
\multicolumn{2}{|c|}{}                                                                                        & Deepfake-3       & 0.78               & 0.86            & 0.82              &                        \\ \hline
\multicolumn{2}{|c|}{\multirow{2}{*}{\textbf{Gboost}}}                                                        & Deepfake-2       & 0.82               & 0.67            & 0.74              & \multirow{2}{*}{76\%}  \\ \cline{3-6}
\multicolumn{2}{|c|}{}                                                                                        & Deepfake-3       & 0.72               & 0.85            & 0.78              &                        \\ \hline
\end{tabular}
\end{table}

\begin{figure}[t!]
    \includegraphics[width=\linewidth]{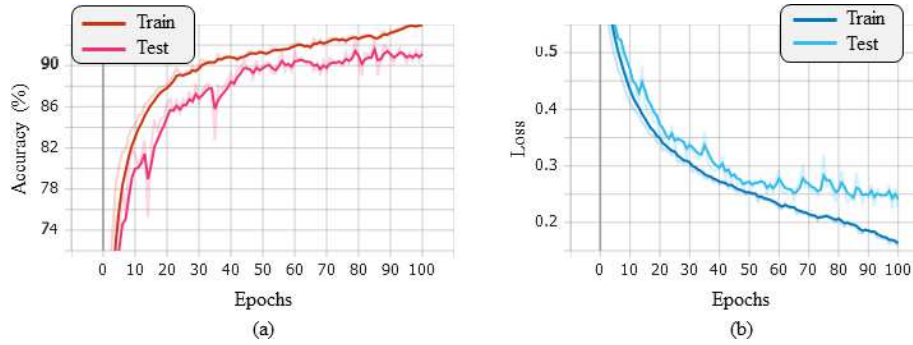}
    \caption{Training accuracy ($\%$) values (a) and loss values (b) were shown in the ordinate axis in (a) and (b) respectively, calculated over 100 epochs (x-axis) of Deepfake-2 and Deepfake-3 class images. The best accuracy was $92.75\%$ with a loss value of $0.23$. 
 }
    \label{fig:res}
\end{figure}

Figure~\ref{fig:res} shows the results obtained considering \textit{Method 2}. We are able to distinguish the images given by the two classes reaching $92.75\%$ as the best classification accuracy value (Loss value equal to $0.23$). 
We are not surprised that we obtain better results than in Method 1. This happens because neural architectures are optimized to infer to their highest degree those features in order to solve the proposed task.
The binary classification task seems to be at first glance a simple task. This is incorrect since RESNET-18 is comparing images of people's faces for both classes and it is well understood that the task that we propose has a higher difficulty than the classic classification tasks (such as classifying with respect to gender). It is clear that 
the neural architecture is not focusing on the semantics of the synthetic data, but is able to capture those intrinsic features that allow us in our case to classify with respect to the number of style-transfer manipulations performed. 
It is important to note another key element in our classification task. As demonstrated by Guarnera et al.~\cite{guarnera2020fighting} each generative architecture leaves a unique fingerprint on the synthetic data. This is the main reason why deepfake detection methods achieve excellent classification results. Under this aspect, our classification task turns out to be more complicated, since we are analyzing images manipulated by the same architecture. Consequently we expect that the images belonging to the two classes contain almost the same fingerprint. Despite this, we manage to solve the proposed task very well. In general, an image manipulated via generative architectures will contain a first unique fingerprint. If the same synthetic content is subjected to a further style-transfer operation (considering the same or other GAN architectures) then the multimedia data will contain another unique fingerprint that we could somehow consider as a ``combination" with the first trace left by the GAN engine. 

Having demonstrated the high accuracy results obtained in the classification task, it may be useful to understand, for complete and accurate forensic ballistics analysis on deepfake images, whether mathematical properties of the style transfer operation are satisfied.
This further analysis is reported in Section~\ref{sec:mat}.

\section{Mathematical Properties}
\label{sec:mat}

Let A, B, and C be three input images. 
We want to demonstrate:
\begin{itemize}
    \item \textit{\textbf{Property 1:}} $A \bigoplus \phi = A$
    \item \textit{\textbf{Property 2:}} $A \bigoplus B = B \bigoplus A$
    \item \textit{\textbf{Property 3:}} $(A \bigoplus B) \bigoplus C = A \bigoplus (B \bigoplus C)$
\end{itemize}
where 
$\phi$ is the neutral element. 

As regards the last two properties we calculated the \textit{Structural Similarity Index Measure} (SSIM) scores and SSIM maps between the output images. In general we expect that all the properties listed above are not satisfied because even minimally, the GAN architecture will manipulate the source image. For this reason, we can also analyze the style transfer with respect to the RGB colors, since the source is also manipulated with the characteristics related to the color distribution of the target. For this reason, we compute two RGB color histograms ($H_1$ and $H_2$)~\footnote{e.g.: supposing we want to prove property 2, then $H_1$ and $H_2$ will be the RGB color histograms of the deepfake images obtained from $A \bigoplus B$ and $B \bigoplus A$ respectively.} related to the two output deepfake images 
and compare them by using~\footnote{\url{https://docs.opencv.org/3.4/d8/dc8/tutorial_histogram_comparison.html}} (i) \textit{Correlation($H_1$, $H_2$)} with an output defined in the interval $[-1;1]$ where 1 is perfect match and -1 is the worst; (ii) \textit{Chi-Square($H_1$, $H_2$)}: with an output defined in the interval $[0;+\infty[$ where 0 is perfect match and mismatch is unbounded; (iii) \textit{Bhattacharyya distance($H_1$, $H_2$)} with an output defined in the interval $[0;1]$ where 0 is perfect match and 1 mismatch.



\paragraph{\textbf{Demonstration 1:}}
We want to try to figure out if there is a neutral element $\phi$ in the style-transfer operation. The neutral element will be that target image such that the style-transfer operation between the source and target gives as output the same source image. 
The neutral element should be a simple image, just a neutral image. Therefore we fix as targets: an all-white image and an all-black image. 
\begin{figure}[t!]
    \includegraphics[width=\linewidth]{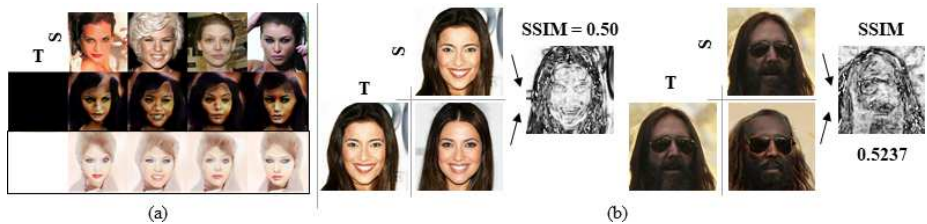}
    \caption{Style transfer operations. S and T represent the source and the target, respectively. (a) The source are images of people's faces and the target are total black and total white images. (b) The source and target are the same images.
 }
    \label{fig:p1}
\end{figure}
As shown in Figure~\ref{fig:p1}(a), the fixed targets seem not to satisfy property 1. The neutral element could be a target image with the same characteristics as the source. For this reason we try to check if it is the source image itself. From the results shown in Figure~\ref{fig:p1}(b), we can say that there is no neutral element because even in the case where target and source represent the same face, the architecture GAN will reconstruct equally the output image, going to transfer, even if equal, characteristics of the target in the source.

\begin{figure}[t!]
    \includegraphics[width=\linewidth]{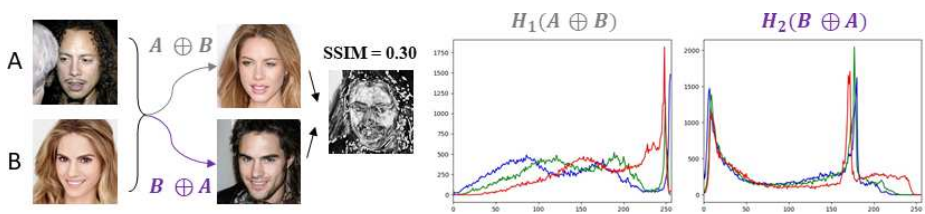}
    \caption{Style transfer operations related to \textit{Property 2}. $H(\cdot)$ represents the function that computes the RGB histogram.
 }
    \label{fig:p2}
\end{figure}

\paragraph{\textbf{Demonstration 2:}} 
Figure~\ref{fig:p2} shows an example of the outputs obtained. The latter turn out to be visually different. Moreover, the 0.3009 score and the SSIM map show us that this property is not satisfied either. 
Finally, we check if the style transfer referred to RGB colors of Property 2 is satisfied. As shown in Figure~\ref{fig:p2}, the distribution of colors is already visually different. We compare the histograms using the metrics listed above. The following values were obtained: \textit{Correlation}: $0.47$; \textit{Chi-Square}: $699.03$; \textit{Bhattacharyya distance}: $0.56$. This shows that even style transfer referring to RGB colors of Property 2 is not satisfied.

\begin{figure}[t!]
    \includegraphics[width=\linewidth]{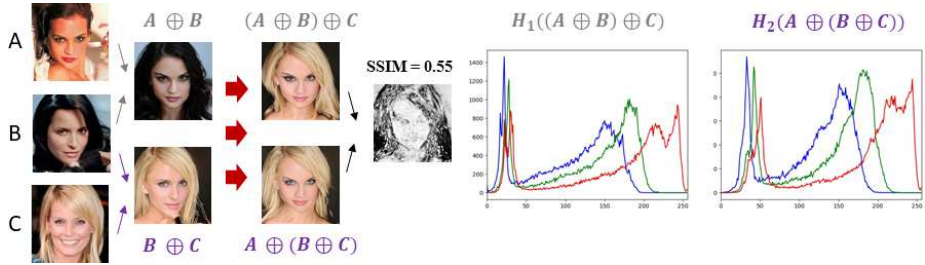}
    \caption{Style transfer operations related to \textit{Property 3}. $H(\cdot)$ represents the function that computes the RGB histogram.
 }
    \label{fig:p3}
\end{figure}

\paragraph{\textbf{Demonstration 3:}}
Figure~\ref{fig:p3} shows an example of the outputs obtained. We can observe that the resulting images are very similar to each other. The score 0.55002 and the SSIM map show us, however, that even this property does not seem to be fully satisfied. Let's check if the style transfer referred to RGB colors of Property 3 is satisfied. As shown in Figure~\ref{fig:p3}, the distribution of colors is visually very similar. The difference values calculated between the two RGB histograms are as follows: \textit{Correlation}: $0.931$; \textit{Chi-Square}: $10.52$; \textit{Bhattacharyya distance}: $0.21$. Considering a set of 1000 randomly selected images, we get on average a \textit{Correlation}: 0.863 (with variance = 0.0034), \textit{Chi-Square}: 14.255 (with variance = 31.2064) and \textit{Bhattacharyya distance}: 0.262 (with variance = 0.0019). From the results obtained we can say that Property 3 is almost satisfied. So the latter can be rewritten in the following way:
    $(A \bigoplus B) \bigoplus C \simeq A \bigoplus (B \bigoplus C)$.

\section{Conclusions}
\label{sec:conclusion}

In this paper, we proposed a first approach to investigate forensic ballistics on deepfake images subjected to style transfer manipulations. In detail, we tried to define if a synthetic image has been subjected to one or two deepfake manipulation processes (performed using StarGAN-v2) by proposing two possible methods: one purely analytical and another based on a deep neural architecture. To better understand the type of style-transfer manipulation we also demonstrated several mathematical properties of the operation itself. From the excellent results obtained, this study represents a good starting point to create more sophisticated methods in the field of forensic ballistics of deepfake images. In future works we will address this issue considering also other different manipulations (e.g. face reenactment, face swap) and we will try to generalize as much as possible also on audio and video deepfakes.

%
%

 \bibliographystyle{splncs04}
 \bibliography{main}
\end{document}